\documentclass[lettersize,journal]{IEEEtran}%,draftcls,onecolumn
\usepackage{amsmath,amsfonts}
\usepackage{algorithmic}
\usepackage{algorithm}
\usepackage{array}
\usepackage[caption=false,font=normalsize,labelfont=sf,textfont=sf]{subfig}
\usepackage{textcomp}
\usepackage{stfloats}
\usepackage{url}

\usepackage{verbatim}
\usepackage{graphicx}
\usepackage{cite}
\usepackage[table]{xcolor}
\begin{document}

\title{A Graph-Based Conceptual Modeling Framework to Design for Human–Robot Coordination}

\author{Martijn~IJtsma,~Salvatore~Hargis
        % <-this % stops a space
\thanks{Martijn IJtsma and Sal Hargis are with the Department
of Integrated Systems Engineering, The Ohio State University, Columbus, OH, 43210 USA (e-mail: ijtsma.1@osu.edu; hargis.29@osu.edu).}
\thanks{Manuscript received June 6, 2024; revised December 17, 2025 and November 3, 2026. This work was supported by an NSF Faculty Early Career Development (CAREER) award under Award Number 2238402.}}

% The paper headers
\markboth{IEEE TRANSACTIONS ON HUMAN-MACHINE SYSTEMS, VOL. 
XX, NO. X, month year}%
{Shell \MakeLowercase{\textit{et al.}}: A Graph-Based Conceptual Modeling Framework to Design for Human-Robot Coordination}

% \IEEEpubid{0000--0000/00\$00.00~\copyright~2021 IEEE}
% Remember, if you use this you must call \IEEEpubidadjcol in the second
% column for its text to clear the IEEEpubid mark.

\maketitle

\begin{abstract}
Studies of human-robot interaction in dynamic and unstructured environments show that as more advanced robotic capabilities are deployed, the need for cooperative competencies to support collaboration with human problem-holders increases. Designing human–robot systems to meet these demands requires an explicit understanding of the work functions and constraints that shape the feasibility of alternative joint work strategies. Yet existing human–robot interaction frameworks either emphasize computational support for real-time execution or rely on static representations for design, offering limited support for reasoning about coordination dynamics during early-stage conceptual design. To address this gap, this article presents a novel computational framework for analyzing joint work strategies in human-robot systems by integrating techniques from functional modeling with graph-theoretic representations. The framework characterizes collective work in terms of the relationships among system functions and the physical and informational structure of the work environment, while explicitly capturing how coordination demands evolve over time. Its use in conceptual design is demonstrated through a case study in disaster robotics, which shows how the framework can be used to support early trade-space exploration of human-robot coordination strategies and to identify cooperative competencies that support flexible management of coordination overhead. These results show how the framework makes coordination demands and their temporal evolution explicit, supporting design-time reasoning about cooperative competency requirements and work demands prior to implementation.
\end{abstract}


\section{Introduction}
\IEEEPARstart{R}{obotic} technologies have the potential to improve efficiency, productivity, and safety in unstructured environments such as disaster response, healthcare, and space operations. However, evidence across human-robot interaction, human factors, and automation research shows that autonomous systems routinely fail to deliver these benefits in real-world field settings \cite{murphy_disaster_2014,Woods2004}. For instance, studies of automation surprise and mode confusion demonstrate that increasing autonomy often adds new demands for supervision and coordination rather than reducing workload \cite{sarter1995world,Wiener1980}. Reviews of human workload in human-robot interaction find that humans must monitor and (re)direct autonomous behaviors, diagnose failures, and share robotic information with other operational actors \cite{Goodrich2007b,chen2014human}. These findings emphasize the need, during system design, to look beyond autonomous capabilities in isolation and to examine how envisioned robot functions interact with the broader social-organizational context in which they must collaborate with human roles.

The computational framework introduced in this paper is intended for conceptual system design, when it can be particularly difficult to understand what cooperative competencies autonomous robots should possess to be useful assets in a mission context. Although interest in robot cooperative competencies is growing, primarily around the framing of human-robot teaming \cite{natarajan_human-robot_2023}, most existing computational frameworks for human-robot interaction formalize collaboration primarily for algorithmic purposes (e.g., task models for real-time robot decision-making such as \cite{rabby_learning-based_2022,Gombolay2017ComputationalPreferences}). In contrast, the few frameworks or approaches that target conceptual system design (e.g., \cite{vicente99,Johnson2014a}) rely on qualitative or static models that do not capture the temporal dynamics of coordination--aspects that computational models are well suited to explore. 

This paper introduces the Joint Strategy Analysis Toolkit (JSAT), a computational framework designed to support conceptual system design by enabling designers to explore how humans and robots coordinate their actions over time and to identify the cooperative competencies robots require before substantial investments are made in physical prototypes. JSAT integrates principles from joint activity theory, graph and network theory, and computational simulation to characterize the structure of collaborative work and to reveal the temporal work demands associated with alternative human-robot strategies. Building on these established concepts, JSAT's primary contribution is methodological: this novel integration conceptualizes management of interdependencies as its own form of work that should be included in the design of system architectures, mapping it across four distinct layers to make coordination overhead an explicit design consideration. This ultimately affords dynamic analytic capabilities for conceptual system design that is unsupported by current static or run-time focused frameworks. While the JSAT framework is intended to be domain agnostic to human-machine teams, the utility of the framework for early-stage design is demonstrated and currently evaluated through a specific case study in disaster robotics.

\section{Background}

\subsection{The Costs of Coordination in Robotic Operations}
Robots offer unique capabilities in unstructured and high-consequence operations, such as aerial mobility, durability to extreme conditions, and the ability to navigate confined or unstable spaces \cite{murphy_disaster_2014}. Unstructured environments are characterized by variability and unpredictability, with work demands that require continuous reassessment and adaptation of work strategies \cite{hutchins1995cognition,casper2003human}. Human-robot operations in these domains constitute a form of joint activity \cite{clark1996using,Klein2004a}, in which humans and robots must coordinate their interdependent activity to achieve shared goals. Central to joint activity is functional interdependence, where the activity of one agent is dependent on that of another, and vice versa, requiring agents to adapt and synchronize their activities with those of others \cite{Klein2004a}. Interdependencies can arise from constraints in the shared work environment, such as access to physical and informational resources, and agents' capabilities, shared goals and roles \cite{Klein2004a}. 

Successful coordination enables the benefits of joint activity, enhancing the system's ability to adapt and achieve shared goals. However, participation in joint activity also imposes coordination costs, as it requires agents to invest cognitive resources in monitoring, projecting, and redirecting other party's activity, while also making one's own actions observable, predictable, and directable to other parties \cite{Klein2004a}. In environments where cognitive resources are limited, humans manage these costs by adjusting coordination strategies: shifting toward more cognitively economical forms when workload is high, and using more resource-intensive but more thorough forms when workload is low \cite{Entin1999}. 

In human-robot coordination, however, asymmetries in cooperative competencies make coordination more rigid and cognitively demanding than in all-human teams \cite{Klein2004a,hoffman2019evaluating,chen2014human}. Autonomous systems often display brittle, literal-minded behavior, leading to failures or the need for intervention when contextually rich situations create mismatches between the robot's programmed logic and the appropriate contextual response \cite{smith97}. Consequently, understanding the coordination requirements and demands of human–robot work is essential for designing robotic systems that better support humans in sustaining effective coordination during complex operations.

\subsection{Design Frameworks for Human-Robot Coordination}
Most existing frameworks that model human-robot collaboration aim to automate and optimize real-time robot control \cite{thomaz2016computational,natarajan_human-robot_2023}. Since these models are intended to be predictive, they are typically applied later in the system development process when technologies are more fully defined and the focus has narrowed to refining a specific technology's behavior. For instance, Rabby et al. \cite{rabby_learning-based_2022} adjusted robot autonomy based on observed performance, while Jin et al. \cite{jin_learning_2023} derived real-time optimal control policies from predictions of human behavior. Similarly, other work has formalized real-time task allocation and scheduling for multi-robot teams \cite{Gombolay2017ComputationalPreferences} and improved collaboration fluency through dynamic task scheduling \cite{hoffman2019evaluating}. While these frameworks demonstrate significant algorithmic advances in enabling real-time robotic behaviors and human-robot fluency bounded, real-time scenarios, they serve a fundamentally different purpose than the analytical frameworks needed during early, conceptual phases of system design, which can be used to question the fundamental ways that automated or human actors can work together to achieve operational goals.

Conceptual system design precedes efforts to model the details of real-time control; its primary goal in this phase is to design the overarching future work practice and imagine how agents can best support one another. Generative (or formative \cite{vicente99}) modeling approaches, such as the JSAT presented here, support this phase by helping designers analytically reason about the implications of their decisions on human-robot collaboration before more granular investments are made into any one technology's development. Rather than claiming predictive validity, the goal is to help designers learn about the problem space of the future system, conceptualize a wide range of system configurations, and gain insight into the coordination demands and trade-offs due to human-robot interdependence. Without generative models, designers must navigate this early phase relying on various frameworks with distinct limitations. Some rely on descriptive guidance from review articles that examine design considerations, such as \cite{simoes2020factors}, which covers physiological, biomechanical, and psychosocial factors for the design of cobots in industrial settings, or hierarchical models of robot autonomy \cite{beer2014toward}. However, these taxonomic approaches have been critiqued outside the human-robot interaction community for oversimplifying the complexity of human-machine interaction \cite{roth2018preface}. Coactive Design \cite{Johnson2014a} offers a structured approach for conceptual reasoning through interdependence analysis; however, it does not explicitly model the work of coordinating itself, nor is it as scalable to multi-human, multi-machine systems. Similarly, distributed artificial intelligence models \cite{jennings1996coordination} formalize agents' beliefs, desires, and intentions to illustrate how aligned mental models enable coherent group behavior. However, this model focuses on the logic of individual agents, rather than representing the work of coordinating and how it can vary over time as demands of the work domain change.

The few frameworks for analytical reasoning about human-robot joint activity face two primary limitations that constrain their design utility: an inability to capture the temporal dynamics of coordination, and a focus on localized agent-to-agent interaction. First, because they rely on static, diagrammatic representations, these frameworks fail to capture how coordination demands can change over time. In reality, work demands change over time, often fluctuating from periods of relative ease to intensive work, necessitating support for multiple types of coordination strategies to deal with a wide array of scenarios one may expect in operations. Without computational models to explore various ways in which agents can sequence and synchronize, designers must often rely on resource-intensive methods such as Wizard-of-Oz studies, in which human experimenters simulate autonomous robot behaviors. Computational models like JSAT help designers map the operational scenarios a system may face directly to the coordination strategies its architecture must support to manage them.

Second, many existing frameworks focus coordination to direct interactions between agents, failing to capture the broader operational system in which they operate, which affects how a designer should consider coordination strategies. Even within a human-robot dyad, coordination is shaped by a web of interdependencies spanning multiple goals, shared resources, functions, and the agents who perform them. Tracing how a localized design change cascades through broader information flows impacts coordination and is an important design consideration, yet remains largely unsupported by existing frameworks focused solely on agent to agent interaction. Graph-based representations offer a path forward by encoding far-reaching interdependencies, making it possible to use graph theory to uncover structural relationships that become more challenging to identify as operations scale.

To address these limitations, we propose JSAT, a computational, graph-based framework that builds upon prior work in computational modeling of work functions for human-robot systems \cite{IJtsma2019a}. JSAT addresses the scalability limitation by formalizing the system as a network, where nodes and edges capture interdependencies among human and robot functions, roles, and resources. To address the temporal limitation, these graph-based models can be dynamically simulated to capture the timing and sequencing element of coordination as work conditions change over time. Ultimately, one of the goals of this computational approach is to improve both the analytical efficiency and scalability of existing generative and formative models, allowing designers to explore how system architectures shape coordination strategies and influence a team's adaptability to changing work demands.

\section{A Graph Representation of Joint Activity}
To support conceptual system design and reveal coordination demands arising from interdependencies, the modeling focuses on work functions rather than the physical or structural-level interactions emphasized in many existing human-robot interaction frameworks. Rather than specifying robotic structures, algorithms, or interaction mechanisms, functional modeling describes what a system does or ought to do to fulfill its intended purpose under varying work conditions \cite{lind2023foundations}. This approach aims to aid designers in reasoning about how early decisions regarding robotic functions and roles may affect coordination demands before committing to costly prototyping or implementation.

JSAT provides a functional map of the action space of a human-robot system and supports analysis of joint human-robot strategies. Unlike traditional modeling approaches, JSAT is not intended for verification or validation. Instead, its purpose is to support exploratory design by allowing designers to represent and reason about interdependencies in envisioned human-robot systems. Accordingly, JSAT is best used iteratively throughout the design process, enabling designers to revise concepts and assess how individual design decisions propagate to system-level consequences. This systems-oriented perspective aims to provide a basis for anticipating how design decisions shape coordination demands and evaluating alternative coordination strategies early in design.

Figure~\ref{fig:graph-of-collab-nav} illustrates the layered structure of the framework (details on nodes and edges appear in Section~IV). The framework represents activity within a work domain as a set of functional nodes coupled through resource nodes, formalized as a directed bipartite graph:

\begin{figure}
    \centering
    \includegraphics[width=0.85\linewidth]{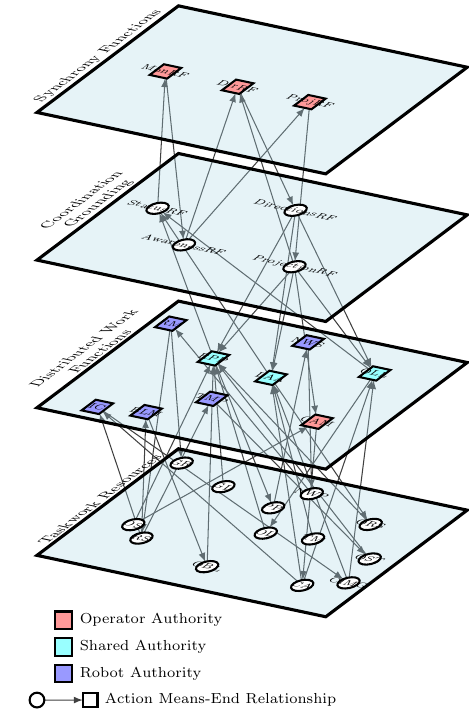}
    \caption{Graph representation of joint activity.}
    \label{fig:graph-of-collab-nav}
\end{figure}

\begin{equation}
    G = (U,V,E), E \subseteq \left\{\{u,v\},\{v,u\} | u \in U, v \in V\right\}
\end{equation}

$V$ is a set of nodes representing tangible and intangible resources for activity, such as information required for performing a function. Collectively, these nodes form the environmental base layer in Figure~\ref{fig:graph-of-collab-nav}, capturing characteristics and constraints of the shared work environment, similar to qualitative descriptions of work domain constraints \cite{Rasmussen1994}. $U$ is a second, disjoint set of nodes representing functions that transform states of the work environment. These nodes constitute the distributed work layer. A function node represents a potential means for action by an agent, where an action is the actualization of a function in context \cite{lind2023foundations}. $E$ is a set of directed edges between the nodes of $U$ and $V$, defining how functions use, modify, or produce resources. These relationships capture how each agent's activity is embedded within the ongoing activity of others: the end of one agent's action may become the means for another agent's action.

The resource and function nodes are organized into four layers to represent environmental and coordination constraints and the functions involved in managing them. By modeling relationships between the layers, the resulting network enables analysis of how dependencies between human and robot functions shape coordination demands. The following subsections describe what aspects of work are captured at each layer; together they form and reflect the theoretical and empirical foundations for the modeling.

\subsection{The Taskwork Resources Layer}
Human and robot work can be described in terms of the information and physical resources in the environment that activities serve to create, control, or maintain. The properties of the work environment--as reflected in these resources--can substantially shape work demands \cite{hutchins1995cognition,vicente99}. Agents act upon and respond to changes in their informational and physical environment \cite{hutchins1995cognition}. Therefore, when analyzing interdependencies between human and robot functions, it is necessary to identify the environmental resources relevant to their activity, along with their temporal, spatial, and functional characteristics. Work demands arise from how these resources change and interact over time and space, including resource limitations, required knowledge states, and the rate at which information or physical conditions change. For instance, in  search and rescue, the tempo and sequencing of activity are constrained by environmental features, such as dense close quarters versus open expanses, which affect how far ahead the human and robot can perceive and plan.

Successful activity in dynamic environments relies on the ability to reason about and manage resources to formulate feasible courses of action \cite{Rasmussen1994,vicente99}. For instance, an activity may require a tool to be available at the correct location and time. Meaningful activity, including coordination with other agents, therefore depends on understanding relationships among resources, independent of the functional means for transforming them.

\subsection{The Distributed Work Layer}
The work required to bring about desired environmental states while managing constraints is modeled as a set of functions on the distributed work layer. These functions represent the means by which activity produces desired effects in the environment. Distributed functions can be modeled independently of what role humans or robots have in realizing the functions \cite{Pritchett2014a} or can model agent-specific competencies as functions of robotic or other technological assets and/or cognitive functions of human agents. Regardless of the specificity and scale of the model, the function nodes should capture what work is required to achieve the objectives and goals of the system. 

In the framework, each agent's role relative to a function can be characterized by three attributes: competency (capable of performing an action), authority (permitted to perform it) and responsibility (accountable for its outcome) \cite{Woods1985}. Thus, through the ascription of roles to agents, function nodes within the network can be assigned to $n$ human and/or robotic agents, creating $n$ subgraphs along each dimension:

\begin{equation} 
    H_n = (U_n,V_n,E_n), U_n \subseteq U, V_n \subseteq V, E_n \subseteq E
\end{equation} 

Each subgraph represents the role of a single agent. The nodes and edges connecting these subgraphs capture interdependencies between roles, where the end (resource) of a function performed by one agent becomes the means (resource) for another agent's function. An effective human-robot system must manage its operating conditions by formulating strategies to bring about desired states in the external environment. This requires understanding how various states are ordinally dependent \cite{Rasmussen1994,vicente99}. The directed graph captures this ordinality by encoding means–end relationships as directed links, making explicit the constraints on how environmental changes can be brought about over time.

Moreover, effective human-robot systems adapt to contextual constraints—such as available time and cognitive resources—by shifting between "contextual control modes" \cite{Hollnagel1993}. Because strategies for long and short time-horizons differ significantly, conceptual models must explicitly anticipate and support this dynamic range of adaptive behaviors during system design.

\subsection{The Coordination Grounding and Synchrony Layers}
An effective human-robot system must manage coordination demands in ways appropriate to its operational context. Well-designed systems should be capable of recognizing when coordination is necessary or beneficial \cite{Johnson2014a} and should possess a knowledge base of transformations that follow functional means-end relationships to achieve coordinated activity. Design should therefore account for how agents navigate organizational constraints and ensure multiple feasible strategies suited to diverse contexts.

Thus, effective system design requires reasoning about taskwork resources and distributed work, but also about the resources and functions that support coordination among agents. The coordination grounding and synchrony layers formalize these aspects, allowing designers to analyze how coordination demands arise from interdependencies among agents and how they may be managed through design interventions. By explicitly representing coordination, the framework enables system designers to anticipate where coordination resources, communication mechanisms, or synchrony functions must be supported in the eventual human–robot system. 

Distributing work among agents offers benefits such as increased efficiency through parallelization, enrichment through complementary perspectives or skills, and robustness through cross-checking and error detection \cite{Klein2004a}. However, such distribution also introduces interdependencies that create cognitive work for information sharing and coordination. These factors, here referred to as organizational constraints, may stem from the system itself or from the broader organization of which it is a part. When the activity of one agent is dependent on the activity of another agent and vice versa, communication and coordination are necessary to synchronize the actions and maintain common ground. 

The framework models the work involved in communication and coordination as additional resources and functions that connect to the primary graph of taskwork resources and distributed work functions:

\begin{equation}
    G_{co} = (U_{co}, V_{co}, E_{co})
\end{equation}

This coordination graph follows the same general nodes and edge definitions as the primary graph, except that nodes have qualifiers distinguishing their coordinative nature. Resource nodes represent coordination resources, such as one agent's awareness of another agent's internal system state (e.g., a human operator's awareness of the robot's battery level). The functional nodes represent the work or overhead of coordinating and synchronizing distributed work, with edges that capture how they interact with and manage coordination resources. Together, these elements constitute the coordination grounding and synchrony layers in Figure~\ref{fig:graph-of-collab-nav}.

Studies and theories on coordination have identified several core requirements for joint activity: observability, directability, and predictability of behavior \cite{Klein2004a}. Functions supporting these requirements may include monitoring and inferring the state, beliefs, and intent of other agents, intervening to redirect their behavior, and diagnosing gaps in common ground \cite{Klein2004a}. In effective human-robot systems, the load for coordination functions is not carried by a single agent \cite{Klein2004a}. For human and robot collaboration, coordination functions must therefore be explicitly designed into robotic capabilities to avoid all coordination overhead falling on human roles. 

In general, distributed work functions represent the activities required to achieve mission goals, whereas synchrony functions represent the activities required to maintain common ground and coordinate between agents. Although these categories can overlap in practice--many functions simultaneously support both mission objectives and coordination--distinguishing them is analytically useful. Without making coordination resources and synchrony functions explicit, the work demands and requirements represented by them risk being overlooked or subsumed in work analysis. The formalization presented here enables designers to deliberately reason about both the ends that must be achieved to maintain synchrony and the means available to support coordination, such as communication protocols, interaction policies, and interface features.

The coordination required in a human–robot system is directly shaped by how competency, authority, and responsibility are distributed and by the interdependence among roles. For example, supervisory control architectures, where humans oversee autonomous robots, impose different coordination requirements than teleoperation architectures, where humans directly control robots \cite{Wiener1980}. Compared to teleoperation, effective supervisory control would have specific requirements for algorithmic capabilities (e.g., transparency) for a human supervisor to understand what (autonomous) capabilities are doing and why, and what robotic systems may do next \cite{sarter1995world}. Within the framework, this relationship is explicit: the roles ascribed to agents at the distributed work layer (e.g., functions assigned to an autonomous agent) shape the resulting topology of the coordination grounding and synchrony layers.

These four layers and the graph formalism provide a framework for designers to reason about how, given envisioned robotic capabilities, the human-robot system could manage resource and work constraints and formulate joint strategies. The next section discusses how this domain-agnostic architecture is instantiated and evaluated within a specific operational design context of a human-robot collaborative navigation task.

%!TEX root = ../bare_jrnl.tex

\section{Using the Framework in System Design}
The remainder of this article demonstrates two primary applications of the framework in the context of human-robot system design. The first involves analyzing the overall structure of the action space of an envisioned human-robot system. This analysis provides insight into the work constraints and interdependencies that human and robot will need to manage during operations. The second focuses on the \textit{temporal} requirements for coordinated activity. Here, computational simulation of the activity graph is used to evaluate the dynamics of work, revealing how interdependent activity will need to be synchronized over time and what operational trade-offs arise in managing coordination overhead.

Both applications are discussed conceptually and then illustrated through a case study in disaster robotics. The case study applies JSAT to analyze human-robot coordination during collaborative navigation between a semi-autonomous unmanned ground vehicle (UGV) and its remote operator.

\subsection{Characterizing the Action Space for Human-Robot Joint Activity}
This subsection describes how JSAT can be used to characterize the topology of a human–robot action space. A significant part of this characterization is the modeling process itself, in which the framework provides scaffolding for a modeler to iteratively explore and identify the work functions and resources necessary for successful performance; thus, insight arises from the modeling activity, not merely from the final model. To construct a JSAT model, resource and function nodes need to be identified from field research, prior research, analytical reasoning, or established operations research models. For example, existing functional modeling approaches suggest that relevant data can be gathered through analysis of process descriptions (e.g., verbal protocols) or through knowledge elicitation techniques such as interviews, observations, and contextual inquiry \cite{roth2008uncovering}. A key part of this process is to continually update the model as the human-robot system being designed becomes more defined during the design process.

Representations can be created for topological analysis to characterize the action space. For instance, analysis of the connections in the graph can reveal where there are means-end relationships that create natural clusters in the action space. As functions, resources, and means-end relationships are added or modified during system design, network metrics enable designers to efficiently explore how such changes affect the overall work structure--for example, by quantifying the relative importance of specific functions and resources based on their dependencies. Two key metrics, modularity and centrality, are discussed to analyze the action space. 

\subsubsection{Modularity}
When representing the ascribed role of agents as modules within a collective joint activity space, modularity (a measure of the strength of the division between modules) captures the relative degree of interdependence between agent roles \cite{IJtsma2019a}. High modularity indicates dense intra-role connections and relatively sparse inter-role connections between agents; thus, an agent's ascribed functions are relatively tightly coupled. In such cases, each agent has a coherent set of coupled functions, suggesting that systems with higher modularity will result in lower coordination demands than those with smaller modularity values. 

Clusters identified through modularity analysis denote functions and resources that are strongly interdependent relative to others. Recognizing these clusters can help system designers understand how human and robotic roles shape the interdependencies that must be managed and, in turn, what coordination mechanisms or synchrony functions may be required.

\subsubsection{Centrality}
Node centrality metrics provide quantitative estimates of the criticality of specific functions and resources to the system's collective performance. In-degree and out-degree centralities identify which functions have the most informational means and ends, respectively. Whereas degree centrality treats all connections equally, eigenvector centrality also accounts for the importance of neighboring nodes, assigning higher scores to nodes connected to other highly central nodes.

For a directed graph like a JSAT model, there are two eigenvector centralities: one based on incoming edges and one on outgoing edges. When computed based on incoming edges, it reflects how much a node depends on influential upstream nodes. For function nodes, this indicates dependence on critical resource nodes and, recursively, on the broader action space. High eigenvector centrality therefore signals strong systemic dependence, implying that the function is sensitive to disruptions in upstream activity. To maintain synchrony, designers should ensure that information resources for the supporting function are updated either from the agent themselves or from another agent coordinating this information with the agent. To achieve synchronized activity, agents require awareness of when and how these functions are performed to appropriately time their own activities. 

When computed on outgoing edges, eigenvector centrality reflects how much a function contributes to, or enables, other important resources and functions. A high centrality score in this direction indicates that the function exerts substantial influence on downstream activities and serves as a key enabler of system performance. To maintain synchrony, designers should ensure that these high-influence functions are performed reliably and on time to support dependent activities under varying conditions. The next section, after the case study, builds on these insights to examine the temporal dynamics of coordinated activity.

\subsection{Case Study: Collaborative Navigation in Disaster Robotics}
Disaster robot operations impose a high cognitive load on responders, and ineffective human-robot interaction remains a major bottleneck to wider deployment \cite{murphy_disaster_2014}. Systems in this domain require teamwork toward shared goals, creating natural interdependencies as humans coordinate with and supervise robotic autonomy. This case study demonstrates how JSAT helps to systematically explore these coordination demands during conceptual design.

% \subsubsection{Characterizing the Base Action Space}
Figure~\ref{fig:graph-of-collab-nav} presents a JSAT model for collaborative navigation. Tables~\ref{tab:functions} and \ref{tab:resources} detail the functional and resource nodes (additional matrices are in the supplementary material). This multi-layered network captures physical and cognitive functions, coupled by resources via directed means-end edges. For example, Imagery Capture (IC) uses Camera Angles (CA) and Robot State (RS) as means to generate Camera Imagery (CIMG).

\begin{table}[tbp]
    \caption{Function nodes and their descriptions by role.}
    \label{tab:functions}
    \centering
    \footnotesize
    \begin{tabular}{lp{2.5cm}p{4cm}}
        \hline
        \bf Lbl & \bf Name & \bf Description \\
        \hline
        \multicolumn{3}{l}{\it Robot Authority} \\
        \hline
        \rowcolor{gray!40}
        IC & Image Capture & Take images of surroundings. \\ 
        RM & Robot Movement & The robot's driving functions (Unicycle model). \\
        NWS & Next Waypoint Selection & Set next drive location (A* output). \\
        BLM & Battery Monitoring & Updates battery level status. \\
        TMP & Temp Monitoring & Monitors robot temperature (Sets OST). \\
        \hline
        \multicolumn{3}{l}{\it Joint Authority} \\
        \hline
        RPP & Robot Path Planning & Determine safe path (A* algorithm). \\
        \rowcolor{gray!40}
        OLL & Obstacle Localize & Update known obstacles map (OL). \\
        \rowcolor{gray!40}
        LAA & Location \& Attitude Assessment & Update precise location/orientation in map. \\
        \hline
        \multicolumn{3}{l}{\it Human Authority} \\
        \hline
        \rowcolor{gray!40}
        CAM & Camera Angle Move & Pointing the sensors of the robot. \\
        \hline \hline
        Mon-RA & Monitor Robot Act. & Tracking current state, behavior, progress. \\ 
        Proj-RA & Project Robot Act. & Anticipating future actions and limits. \\ 
        Dir-RA & Direct Robot Act. & Providing commands/goals to guide robot. \\
        \hline
    \end{tabular}
    \vspace{2pt}
    \parbox{\linewidth}{\scriptsize \textit{Note:} Gray/white rows denote the two optimal graph clusters. Unless specified, functions were modeled solely via means-end relationships to denote a function was performed (no transformation).}
\end{table}

\begin{table}[tbp]
    \caption{Resource nodes and their descriptions.}
    \label{tab:resources}
    \centering
    \footnotesize
    \begin{tabular}{llp{4cm}}
        \hline
        \bf Label & \bf Name & \bf Description \\
        \hline
        RS & Robot State/Orient & Location and current direction. \\
        GR & Goal Reached & End state (terminates simulation). \\
        NWP & Next Waypoint & Next drive point (from A* steps). \\
        GL & Goal Location & End location (A* initial condition). \\
        TM & Terrain Map & 2-D map (Nuclear disaster model). \\
        \rowcolor{gray!40}
        ORS & Observed State & Assessed location and direction. \\
        \rowcolor{gray!40}
        PP & Planned Path & Path followed (A* output). \\
        \rowcolor{gray!40}
        CIMG & Camera Imagery & Images and video feed. \\
        \rowcolor{gray!40}
        OL & Obstacle Locations & Locations of obstacles on TM. \\
        \rowcolor{gray!40}
        CA & Camera Angles & Angle of the robot's camera. \\
        \rowcolor{gray!40}
        OBL & Observed Battery & Assessed battery levels. \\
        \rowcolor{gray!40}
        OST & Observed Temps & Assessed subsystem temperatures. \\
        \rowcolor{gray!40}
        OS & Obstacle Size & Size of obstacles (from TM). \\
        \hline \hline
        \rowcolor{gray!40}
        RAS & Robot Activity Status & Visible activity information. \\
        \rowcolor{gray!40}
        DRA & Directions for Act. & Commands to steer robot activity. \\
        ARA & Awareness of Act. & Human awareness of robot state. \\
        PRA & Projections of Act. & Projection supporting human timing. \\ \hline
    \end{tabular}
    \vspace{2pt}
    \parbox{\linewidth}{\scriptsize \textit{Note:} Gray rows indicate information shared between human and robot. Only means-end relationships were modeled.}
\end{table}

To construct this model, patterns of UGV functions and resources were identified from field studies \cite{murphy_disaster_2014,Stubbs2007} and analytic reasoning \cite{Rasmussen1994}. Roles were defined by competency and authority (Table~\ref{tab:functions}). While alternative work distributions can be explored, this study focuses on one configuration: the robot holds authority over primary driving (RM, NWS), system monitoring (TMP, BLM), and image capture (IC), while the human controls camera angles (CAM). Authority over obstacle localization (OLL), location/attitude assessment (LAA), and path planning (RPP) is shared, shifting between human and robot based on context (e.g., unexpected obstacles). Supporting these dynamic shifts in authority relies on the gray highlighted shared information resources in Table~\ref{tab:resources}.

Modularity and centrality were then applied to analyze the topology of the action space. First, clusters were identified based on what functional and resource nodes are most closely linked through means-end relationships. The gray and white in Table~\ref{tab:functions} indicate the two main clusters. Consequently, when both the human and the robot are ascribed authority over functions belonging to the same cluster, coordination demands are expected to be higher than when the functions in that cluster are ascribed to only one agent. If the human and robot roles were ascribed to perfectly align with the clusters, their roles would have a modularity of 0.285. Because the ascribed human and robot roles do not align perfectly, and part of the functions have shared authority, the best possible modularity that the human and robot can have is 0.082 (and thus the lowest expected coordination demands), when the human performs Location and Attitude Assessment (LAA) and Obstacle Localization (OLL) or performs all three of the shared-authority functions (akin to a more manual form of control). The worst possible modularity is -0.163 (and thus the highest expected coordination demands), when the human performs just Robot Path Planning (RPP), and the robot does Obstacle Localization (OLL) and Location and Attitude Assessment (LAA). Therefore, strategies in which the human plays a significant role in path planning but not the other two shared functions will likely have a higher coordination load.

\begin{figure}
    \centering
    \includegraphics[width=0.43\textwidth]{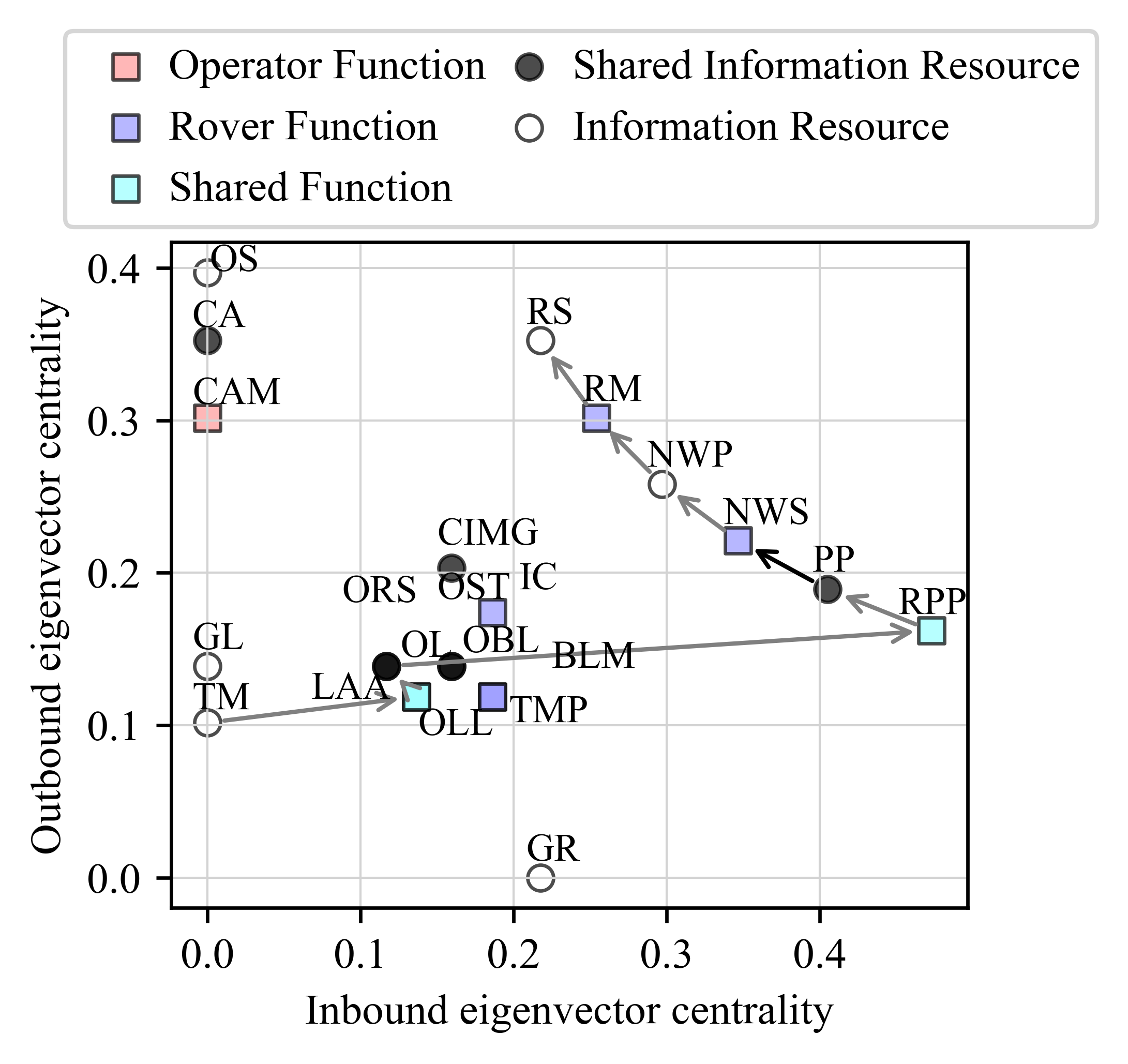}
    \caption{Eigenvector centrality of the function and resource nodes}
    \label{fig:concentric view}
\end{figure}

Second, eigenvector centrality was used to quantify the degree to which each function depends, or is depended upon by other functions and resources. Figure~\ref{fig:concentric view} visualizes eigenvector centrality for both function and resource nodes. The horizontal and vertical axes represent centrality based on incoming edges (how much a node depends on others) and outgoing edges (how much other nodes depend on it), respectively. Nodes located on the right half of the graph depend on many other elements, while those in the upper half are relied upon by many others. Robot Movement (RM) and Next Waypoint Selection (NWS) are both highly central functions, as they depend on and support numerous other functions. Another critical function is Robot Path Planning (RPP), which exhibits high dependence on other nodes.

From this analysis, a designer can infer that synchrony and coordination will be particularly critical among these highly central functions and their immediate neighbors. Specifically, since the robot holds authority over Robot Movement (RM) and Next Waypoint Selection (NWS) but potentially relies on the operator's Robot Path Planning (RPP), synchronizing robotic movement with path-planning processes is likely to represent a major source of coordination load. This relationship warrants careful consideration in subsequent system design. Similarly, the high centrality of several resources--Robot State and Orientation (RS), the Next Waypoint (NWP), and the Planned Path (PP)--suggest that maintaining these resources as accurate, up-to-date, and salient will be crucial for effective coordination.

Building on these insights, the designer can reason functionally about resources and synchrony functions needed to coordinate distributed human-robot work, ultimately identifying functional coordination requirements as a set of coordination resources and synchrony functions. In this case, the human and robot share authority over driving functions: while the robot executes low-level control, the operator can contribute contextual awareness unavailable to the robot, and needs to be able to monitor and redirect the robot’s behavior as needed. Three synchrony functions are highlighted here, corresponding one-to-one to three fundamental requirements of joint activity \cite{Klein2004a}:

\begin{itemize}
    \item \textit{Monitoring Robot Activity}. To coordinate human-robot activities and manage shared control, the human must stay aware of the robot's activity. This requires that robot activities be observable. The model includes a resource Robot Activity Status (representing visible activity information) and a monitoring function (representing the human’s monitoring effort). Edges connect robot functions to the status resource, the status resource to monitoring, and monitoring to a resource Awareness of Robot Activity.
    \item \textit{Projecting Robot Activity}. This function captures the cognitive work for the human operator to predict robotic behavior and synchronize their actions accordingly. A function Projecting Robot Activity (Proj-RA) is a means to generating a projection resource that supports the timing of human actions.
    \item \textit{Redirecting Robot Activity}. This represents the ability to steer the robot's activity for the functions for which human and robot have shared authority. This requires that robot functions be sufficiently redirectable.
\end{itemize}

Further modeling could incorporate additional coordination and synchronization mechanisms; however, for the present demonstration, we restrict the analysis to the  coordination requirements for the functions that the human and robot have shared authority over.

\subsection{Temporal Analysis of Human-Robot Synchrony}
The graph model can also help designers conduct fast-time computational simulations of work dynamics and identify temporal requirements for synchronizing activities in a human-robot system. The graph model formalism can be converted to the computational structures of Work Models that Compute (WMC) \cite{Pritchett2014a,IJtsma2019a}, which can simulate the interactions between work functions over time. A WMC simulation run actualizes the modeled functions by instantiating them as a sequence of actions performed by specific agents, at specific times, and under the demands imposed by a scenario, which grounds the otherwise general graph model in concrete events and constraints. Each simulation run produces a concrete action sequence--essentially as a path through the graph--that reflects one feasible way the human-robot system could perform and synchronize work.

Strategies are defined as categories of such action sequences. Rather than identifying only a specific action sequence for a singular situation, a strategies analysis focuses on identifying and examining classes of activity sequences that share characteristics and are reflective of recurrent patterns of behavior \cite{vicente99}. Strategies can differ in what work is performed, when it is performed, by whom, and how functions are actualized. They also reflect different values and priorities in joint activity, such as minimizing actions and time to achieve a desired outcome, modulating the degree of collaboration between the robot and human to avoid operator workload saturation, and/or maximizing the human-robot system's capacity for adaptation. 

The specific strategy that the system enacts is a product of human and robot actors selecting actions in the moment, based on the evolving context, such as the status of resources (e.g., ``how long ago was this information revised?'', ``how quickly does this information typically change?''), the resource needs of oneself or others for the performance of functions (e.g., ``how up-to-date do I or others need these information resources to be?''), and the current contextual demands (e.g., ``how much time do I have to respond?'', ``how many  cognitive resources do I have available to act?''). These questions illustrate the kinds of considerations that matter when human and robot actors attempt to synchronize their activities.

Thus, the goal of a strategies analysis is to explore how--given dependencies in the work--activities \textit{could} (as opposed to \textit{should}) be synchronized in a variety of feasible ways to meet different contextual demands. Even though actions are ultimately selected by humans and robots during operations, an exploratory analysis of strategies during design can illuminate the operational and cognitive trade-offs that they will need to navigate and the range of ways their activities could be coordinated. Knowledge of this range of feasible activity is important for designing to support strategies, especially around system performance boundaries.

In the JSAT framework, edges represent dependencies within the human and robot activities that must be managed over time in order to achieve shared goals (e.g., one cannot go straight from A to C, but needs to go through B). To identify and examine alternate strategies, edges can be annotated with attributes that indicate the standards or criteria for managing the dependency. For example, edges between functions and resources can be given an attribute that represents how up-to-date an information resource must be for the performance of the associated function, reflecting the degree to which information must remain adequate or current for use. Likewise, attributes on edges going from functions to resources can reflect the actual information currency of the function at some point in time. Other examples of attributes could be latency, reliability, or error rates. 

Among the possible edge attributes, one that is particularly important for coordinating dependent functions is the required up-to-dateness of an information resource relative to a function, here referred to as the desired information currency. The required information currency depends on the characteristics of the resource and the function that uses it, as well as context in which the function is performed. For example, the required currency can be low when information contained in the resource has a long lifespan relative to the operational context (for example, in relatively stable work environments, one party may front-load information in a way that increases its lifespan for another function, such as an operator inputting multiple waypoints ahead for a robot to navigate so that the robot can operate autonomously for a longer period of time). Likewise, required currency can be low when the function is robust to staleness in the resource. 

In the WMC engine, functions are scheduled based on the means-end relationships and their associated attributes. In particular, if a scheduled function uses an information resource that is outside its required currency, its associated upstream function is automatically called to update the resources first. This happens recursively, backtracking through the graph network until all resources are sufficiently up-to-date. More details on the implementation of the simulation engine can be found in \cite{IJtsma2019a} and \cite{Pritchett2014a}. Alternative strategies can then be identified and simulated by varying the required information currency on one or multiple edges of interest.

Beyond scheduling dependencies, the simulation also models how functions operate within and respond to a dynamic environment. By computationally modeling the evolving values of information resources and the functions as mathematical transformations that act upon and change those values, the model can simulate how actions interact with a dynamic environment and with each other (as part of an actualized strategy). For example, the simulation can trace how human and robot actions unfold over time and interact with the dynamics of a UGV and the potentially cascading dynamics of an unfolding situation (e.g., wildfire dynamics). This can reveal how the timing of human and robot activity can create (un)desired system behavior and performance outcomes, and it can identify the conditions under which different levels of information currency succeed or fail, providing a basis for assessing the feasibility of alternative strategies.

Fast-time simulation can therefore be used to explore the feasibility and implications of different strategies. Stricter constraints require functions to be performed more frequently to ensure resources remain sufficiently current, increasing demands on the agents that perform those functions. Relaxing constraints reduces these demands, but also decreases the currency of information resources that downstream functions use, potentially leading to asynchrony, stale information, and reduced system performance. This reflects a balance of synchronization: maintaining sufficient information currency without imposing excessive work demands to keep information up-to-date. By varying temporal constraints, a strategies analysis can reveal how a human-robot system may manage this balance and what minimum standards or criteria for managing dependencies are necessary in a given work environment. 

\subsection{Case Study Continued: Analyzing Temporal Dynamics of Collaborative Navigation}
As Murphy documents \cite{murphy_disaster_2014}, disaster environments impose strict and highly variable sensing constraints: dust, smoke, darkness, and obstacles often limit how far ahead operator and robot can reliably sense or plan. These constraints directly affect the high centrality functions identified in the first part of the case study--RM, NWS, and RPP--which depend on maintaining sufficiently current information in resources resources such as RS, NWP, and PP. For this reason, the temporal analysis focuses on look-ahead distance for sensing and planning and examines how it shapes the range of feasible strategies through which operators and robots can synchronize their activities.

Most disaster robotics teams operate close to saturation as work demands and uncertainty are inherently high. Cognitive and temporal pressures mean that some functions are performed only as often as needed to keep their associated information sufficiently current. For example, an operator may not need to re-check the resource Obstacle Locations (OL) each time the function Robot Path Planning (RPP) is performed if the environment has remained relatively unchanged. By varying currency requirements on the edges in the graph model, the analysis explores different strategies for how a human and robot could synchronize their activities, revealing the minimum information requirements necessary for maintaining synchrony across their shared activities.

The analysis used a simple unicycle model to simulate UGV motion in a debris map adapted from \cite{chen2019detection}, generated by a physics-based simulation of a severe earthquake at a nuclear power plant. Obstacle locations (debris-scattered areas) were treated as unsafe driving zones, with all other remaining areas considered safe. High-centrality function and resource nodes were implemented computationally as mathematical transformations that interact directly with the rover dynamics (see the last columns of Tables I and II). Obstacles awareness was represented as a two-dimensional map updated by the Obstacle Localization (OLL) function. The Plan Rover Path (RPP) function was implemented using an A* algorithm to approximate the optimal path based on the locally available obstacle information. Both OL and RPP incorporated a variable Look-Ahead Distance (LAD) to examine the effects of look-ahead on coordination demands. Functions with lower centrality were modeled at lower fidelity, though the model can be readily extended as the focus of the analysis shifts.

Function timing during runtime was based on the graph model in Figure~\ref{fig:graph-of-collab-nav} and the specified currency requirements for the edges. The primary control logic directs the robot to drive toward and stop at the next waypoint through the function Robot Movement (RM). Once at the waypoint, the function Next Waypoint Selection (NWS) is triggered. Subsequent functions are then scheduled according to their means-end relationships and the currency of the information resources they depend on. For example, if the resource Planned Path (PP) was last updated 25 seconds ago and the currency requirement on the PP $\rightarrow$ NWS edge is 20 seconds, the simulation engine automatically schedules the function Robot Path Planning (RPP) to update PP before performing NWS. Performing RPP may require other upstream functions to be performed first, such as Obstacle Localization (OLL) to first update the resource Obstacle Locations (OL). For this analysis, it was assumed that whenever the simulation schedules a function, the agent having authority over that function performs it instantly without delay or mistakes.

For each simulation run, the currency constraints on the PP $\rightarrow$ NWS edge and the look-ahead distance of RPP and OLL were systematically varied to represent alternate coordination strategies. In practice, operators adjust look-ahead distance and the required currency of information on the fly, based on what they can perceive and how quickly conditions change. However, these parameters were held constant within each simulation run to isolate their individual and joint effects. This allows the simulation to identify the minimal information currency required for short look-ahead distances and how those requirements change as look-ahead increases. Early in conceptual design, such analysis helps map the trade-space and delineate the boundaries of stable coordination that a skilled operator would need to navigate dynamically during actual operations. Required information currency ranged from 0 to 70 seconds in 5-second increments (0 seconds denotes the strictest requirement, where the PP resource must always be updated before performing NWS). Table~I in the supplementary material shows the currency requirements for other edges, which were held constant for all simulation runs. Tested look-ahead distances ranged from 2.5 to 20 meters with variable intervals selected for resolution. In total, 15 currency levels x 9 look-ahead distances yielded 135 simulation runs.
 
WMC records all instances in which one agent performed a function whose incoming information resources were last set by another agent, providing a metric for the required information exchange--a proxy for communication load. The simulation also logged events in which the robot entered an unsafe zone. These events occur when look-ahead distance and information currency requirements are misaligned, causing NWS to act on outdated path information. While real systems include obstacle-avoidance fail-safes, here unsafe-zone entry serves solely as a diagnostic of synchrony breakdown and delineates the feasible envelope of coordination strategies.

\begin{figure}[ht]
    \centering
    \begin{tabular}{c@{}}
        \includegraphics[width=.4\textwidth]{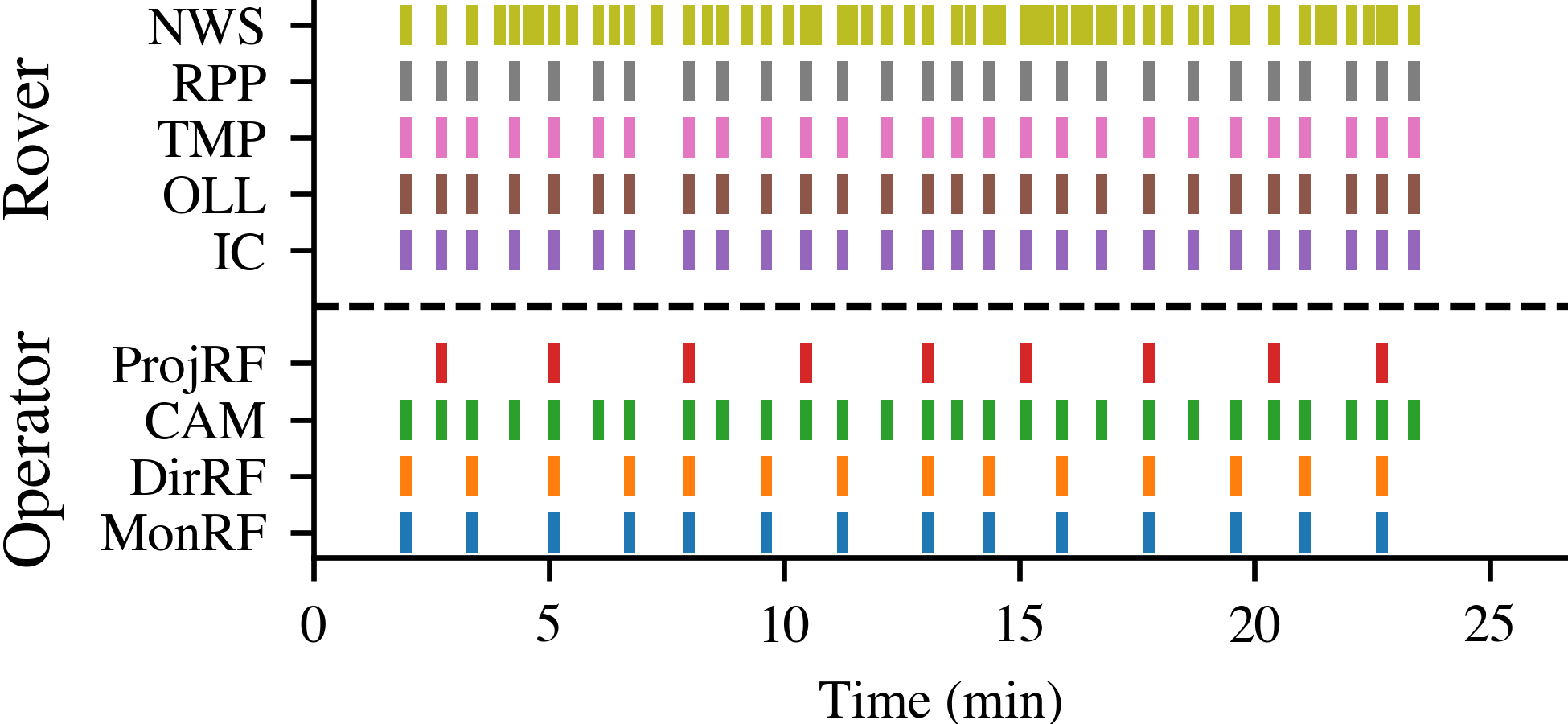} \\
        \footnotesize (a) Look-ahead distance of 6.0 m and PP$\rightarrow$NWS required currency of 35 s. \\
        \includegraphics[width=.4\textwidth]{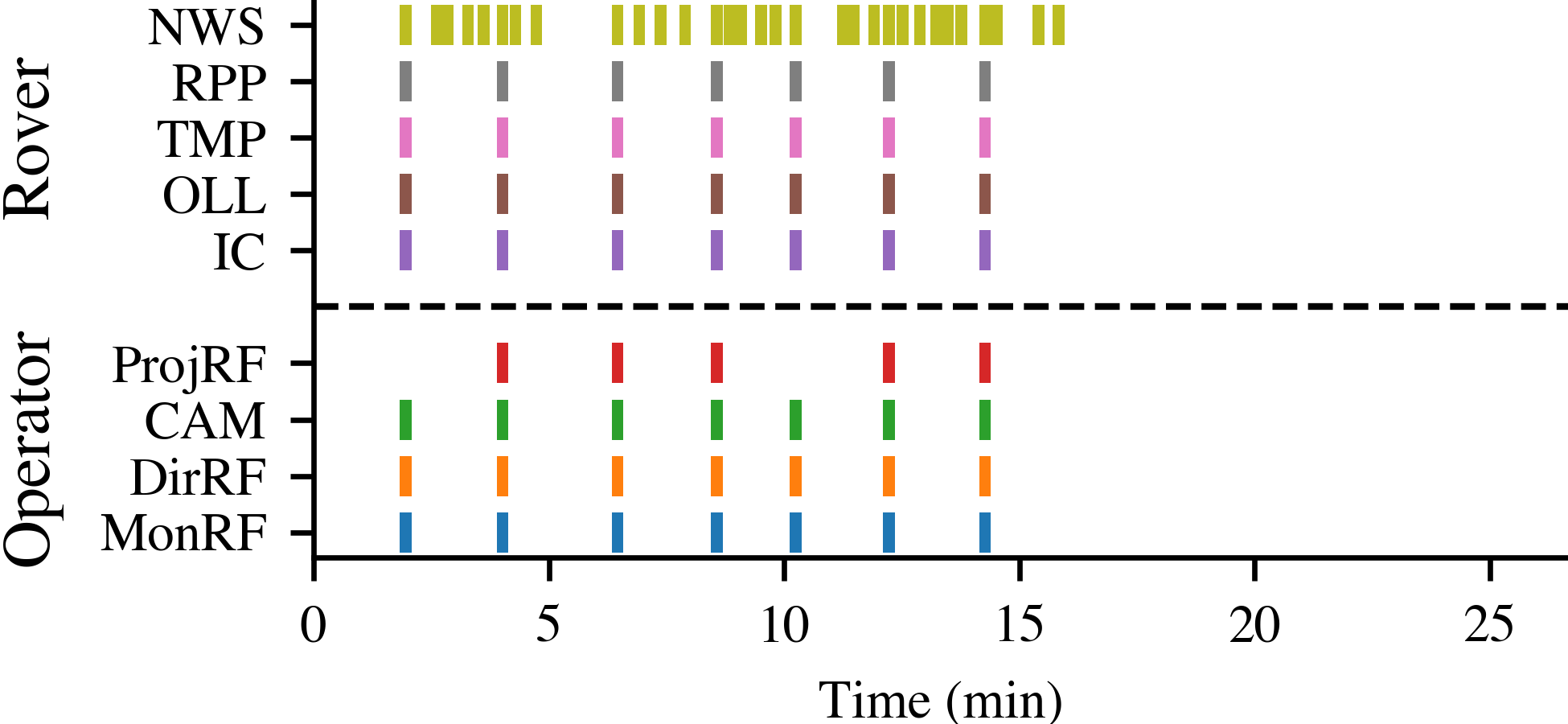} \\
        \footnotesize (b) Look-ahead distance of 15.0 m and PP$\rightarrow$NWS required currency of 100 s. \\
    \end{tabular}
    \caption{Activity traces from the computational simulation for two different strategy parameterizations.}
    \label{fig:activity}
\end{figure}

Figure~\ref{fig:activity} shows the action traces resulting from two simulation runs that differ in currency requirements and look-ahead distance. Each row corresponds to a particular function, and each stripe marks a single occurrence of that function. The figure illustrates how with further look-ahead distances (subfigure (b)), the robot can travel longer between waypoints, as indicated by the increased interval between consecutive occurrences of NWS. It also highlights the recursive effects of functional dependencies: when the currency requirement for PP $\rightarrow$ NWS is relaxed from 35 s (subfigure (a)) to 100 s (subfigure (b)), not only is RPP invoked less frequently, but the upstream functions on which RPP depends, such as CAM or DirRF, also occur less often. While subfigure (a) shows a much denser action trace, suggesting a higher operator workload, further relaxing the currency requirement (allowing PP to be older than 35 s) would cause the rover to enter unsafe zones because the PP information resource becomes too stale relative to the rover's motion. This lower boundary on information currency will be examined further in the following paragraphs.

\begin{figure}
    \centering
    \includegraphics[width=0.4\textwidth]{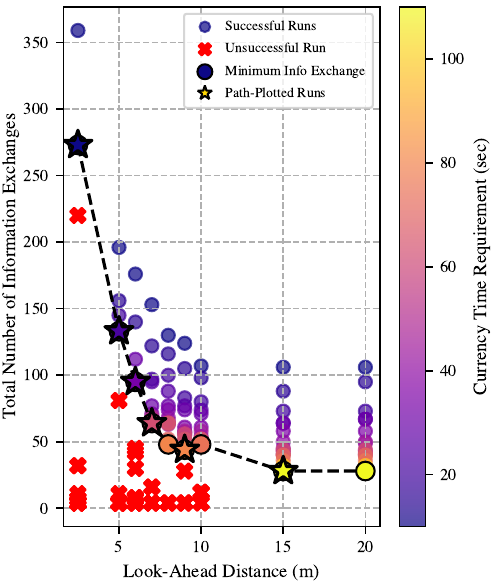}
    \caption{Total information exchanges versus look-ahead distance, where color indicates the currency requirement and the dashed line denotes the minimum information exchange required to satisfy that run's coordination constraints.}
    \label{fig:minimuminfo}
\end{figure}

Figure~\ref{fig:minimuminfo} shows the total number of information exchanges for each simulation run. Each dot represents one run with a unique combination of look-ahead distance and required information currency. These two parameters jointly determine the information exchange required between the human and the robot. Stricter currency requirements increase the required number of exchanges but maintain higher information currency. Meanwhile, the look-ahead distance for RPP and OLL determines how far the rover can travel before the planned path (PP) becomes stale. Larger look-ahead distances permit a more relaxed currency requirement on PP, thereby reducing the number of required RPP updates.

The dashed line in Figure~\ref{fig:minimuminfo} denotes the minimum necessary information exchange as a function of look-ahead distance. Strategies below this line represent ``under-coordination,'' in which  stale and outdated information causes the rover to enter unsafe regions--an indication of synchrony breakdown. In contrast, strategies above the line represent ``over-coordination,'' achieving adequate performance but at a higher-than-necessary coordination cost. This theoretical line therefore highlights a trade-off boundary among the feasible strategies for the human–robot system. Strategies along the boundary avoid updates that are too infrequent to sustain synchronized performance, while also avoiding unnecessarily high work demands that provide little additional performance benefit. Rather than identifying a single ``optimal'' strategy, the line makes this trade-off explicit and uses the generative model to reveal a range of feasible activities that system designers should consider supporting.

\begin{figure}
    \centering
    \includegraphics[width=0.5\textwidth]{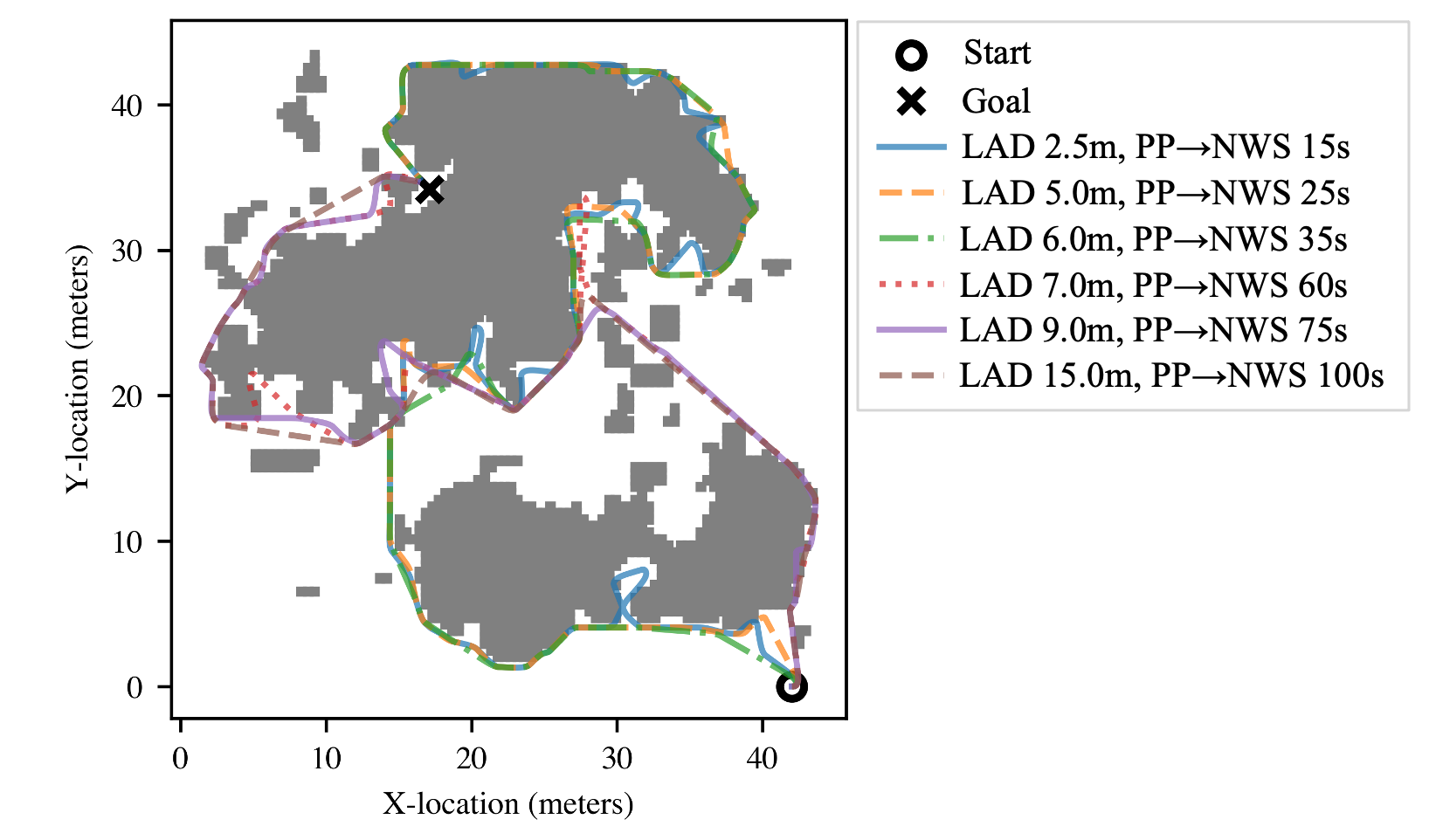}
    \caption{Trajectories of UGV through a nuclear reactor debris field, adapted from \cite{chen2019detection}.}
    \label{fig:trajectories}
\end{figure}

For the simulation runs highlighted with stars in Figure~\ref{fig:minimuminfo}, the resulting rover trajectories are shown in Figure~\ref{fig:trajectories}. These examples illustrate how varying the look-ahead distance (while maximally relaxing currency requirements) affects rover trajectory, distance traveled, and mission completion time. They also demonstrate the impact on operator work demands, as previously shown in Figure~\ref{fig:activity} (which provides action traces for two of the starred runs).

In summary, the strategies analysis quantified a key operational trade-off that a human operator must manage when collaborating with a robot: maintaining sufficiently current information to avoid performance breakdowns while keeping coordination demands tractable. The results indicate two general approaches. One is to plan with a longer looking-ahead distance, which may require greater cognitive effort during each planning episode but reduces the need for subsequent synchrony and coordination as the rover executes the plan. The other is to plan with a shorter look-ahead distance, which lowers the cognitive burden per planning instance but increases the amount of information exchange and frequency with which synchrony must be maintained. 

This balance was revealed through the modeling and simulation of the functional model. The findings suggest several implications for the future design of this human-robot system: (1) Support for extended look-ahead during path planning is a key enabler for flexibly managing coordination cost, as it provides a wider range of strategies for maintaining effective human-robot synchrony; (2) designers should consider hardware, algorithmic, and interface capabilities that expand the feasible look-ahead distance, while also providing support for conditions in which look-ahead distance is constrained (e.g., low visibility or degraded sensing), as such situations impose particularly high coordination demands; and (3) designers should include mechanisms to help operators develop a real-time understanding of appropriate planning distance and update frequencies for the current operational context--for example, displays or cues that convey information currency, plan lifespan, or projected coordination effort. 

\section{Discussion and Conclusion}
While it is well-established that robots should be designed for joint activity with humans \cite{Johnson2014a,Klein2004a,Woods2004}, there are currently very few frameworks that provide design support. Current guidance and frameworks for early-stage design \cite{Johnson2014a,beer2014toward,simoes2020factors} do not explicitly capture the dynamics of human-robot joint activity and are diagrammatic and therefore difficult to scale. This paper presented a computational and representational tool to help designers reason holistically about joint activity and coordination in human-robot systems before those systems are built. This form of reasoning is not supported by prior tools: the formal integration of functional modeling and joint activity theory with graph representations (1) explicitly encodes interdependencies and temporal constraints that require coordination and (2) the explicit modeling of coordination work (grounding and synchrony layers) alongside taskwork treats coordination as a primary design object. Finally, the coupling of this representation with simulation (WMC) enables analysis of dynamic coordination strategies during conceptual design, which existing frameworks do not support. 

The case study demonstrates the analytical capability by walking through the specification of functions, assignment of roles and analysis of the graph topology. The framework operates by progressively specifying work functions, resource dependencies, human-robot roles, and the grounding resources needed for synchrony. Through this process, the case study highlights how coordination demands are driven by both work constraints and human-robot system design. Because this modeling process requires the designer to explicitly articulate interdependencies, it provides a structured way to examine well-known determinants of human-robot system performance, such as observability, predictability, and directability \cite{Klein2004a,Woods2004}.  The framework is not intended to directly reduce workload, but aims to aid designers in anticipating and managing factors that influence workload in HRI. 

Furthermore, the case study demonstrates how the framework uses fast-time simulation to reveal the impact of varying two work and environmental constraints (required information currency and look-ahead distance) on the temporal feasibility and cost of alternative coordination strategies. This illustrates how the framework affords analysis of operational trade-offs around coordination and dimensions along which humans and robots may adapt their coordination behaviors. 
Rather than predict what coordination strategies \textit{will} occur, the model identifies which strategies are possible and at what cost \textit{under specified work and environmental constraints}. The aim is not to describe or predict observed behavior but to provide valid design-time inferences about how work and environmental constraints affect coordination demands.

Reflecting this aim, the simulation deliberately models human and robot agents as perfectly and instantaneously executing their actions to isolate the effects of work and environmental constraints on coordination demands and feasibility of strategies. This minimal modeling of human and robot performance characteristics enables unambiguous attribution of observed trade-offs to structural properties of the activity space (and the human-robot system design) rather than assumed limits of human or machine performance. In other words, if a coordination strategy is challenging under perfect agent execution, a designer can expect it to be definitely challenging in reality. This is helpful in early systems design, when a key goal is to understand the impact of design decisions related to the human-robot system architecture on coordination demands. 

While this case study provides a single snapshot of the models at one point in time, the framework's true value lies in how these models are iteratively developed, refined, and updated to reflect a deepening understanding of coordination demands throughout the design process. Additional performance constraints can be added to the baseline model later in the design process. For example, a logical extension of the case study could model errors, delays, or rigidity in robot behaviors to identify how those constraints could create additional coordination demands and impact the feasibility of strategies. However, for models to be useful for design, we reiterate that there is value in minimal constraint-based modeling initially to help isolate causal mechanisms. 

The use of modeling and simulation for exploratory and diagnostic purposes also warrants a brief discussion of the connection to empirical testing and evaluation during design time.  The framework is not meant to be a substitute for testing and evaluation of human-robot system designs, but to be used in concert with it to offer complementary insights into coordination demands. Computational modeling offers low-cost insight into the dynamics of envisioned systems as new agents, resources, and environmental factors are added to the model. Insights derived from computational analysis can guide further empirical studies or evaluations of proposed designs. 

% A limitations paragraph?
The current paper has limitations that constitute directions for future work. The use of JSAT is presently demonstrated in a single context: collaborative navigation between a human operator and a UGV in disaster response. While the graph modeling structure is designed to be domain-agnostic with the aspiration of applying it to a range of contexts, future work must prove broader generalizability by applying the framework to other human-machine teaming domains. Future validation is also necessary to further evaluate the framework's ability to support design and ultimately create higher-performing human-robot system designs. For example, controlled user studies could compare design decisions made with and without JSAT, ideally in the context of real-world HRI design challenges.
% Last, to validate the scalability of this approach in a design context, the approach needs to be applied to larger-scale human-robot systems. 
% should be benchmarked against existing design frameworks %Need to check whether this makes sense to say--is "benchmarking" used in this way or only in comparing computational frameworks?

Another important direction for future work is to apply the framework to operations involving more than two agents. The framework does not impose assumptions on the number of agents involved in executing functions, and thus can be used to analyze human-robot collaboration at a variety of scales and across scales (e.g., starting with a dyad but then expanding the scale to systems of multiple dyads). This opens the door for envisioning coordination demands and identifying design requirements for scenarios with multiple human team members and robotic assets such as disaster relief, where multiple UAVs inform UGVs and human rescuers about points of interest and obstacles and human rescuers interact with UAV operators to direct their priorities and exchange mission relevant information \cite{murphy_disaster_2014}. As scale increases, so too do interdependencies among agents, underscoring the value of computational and graph-based analysis tools to examine distributed work systems whose complexity makes manual tracing of interdependencies increasingly challenging. 

\bibliographystyle{IEEEtran}
\bibliography{IEEEabrv,references}
 % argument is your BibTeX string definitions and bibliography database(s)
%\bibliography{IEEEabrv,../bib/paper}
%

\end{document}

\title{Supplementary Material -- A Graph-Based Conceptual Modeling Framework to Design for Human–Robot Coordination}

\author{Martijn~IJtsma,~Salvatore~Hargis}

% The paper headers
% \markboth{Journal of \LaTeX\ Class Files,~Vol.~14, No.~8, August~2021}%
% {Shell \MakeLowercase{\textit{et al.}}: A Sample Article Using IEEEtran.cls for IEEE Journals}

\maketitle

{\appendices

\section{Complete Model with Adjacency Matrices}
Figure 1 in the main document presents the complete network. For improved visibility of the edges, a two-dimensional representation is provided in Figure~\ref{fig:full-network} of this appendix. The corresponding adjacency matrices are listed in Tables~I and~II. 

\begin{figure*}[!h]
    \centering
    \includegraphics[width=.8\linewidth]{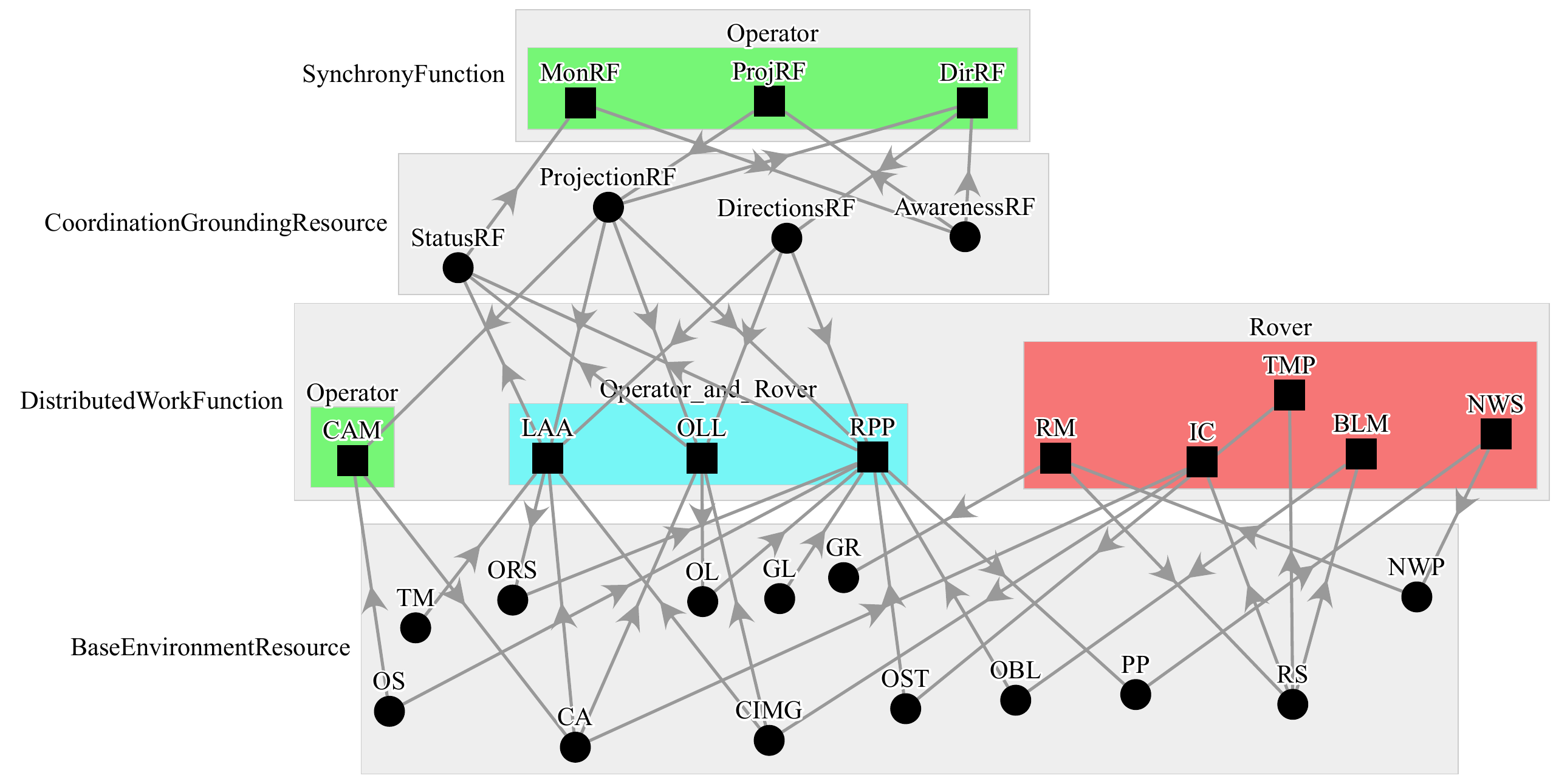}
    \caption{Two-dimensional representation of the complete work model to improve visibility of edges.}
    \label{fig:full-network}
\end{figure*}

Table~I shows the directed relationships from resource nodes to function nodes. The numerical values in the matrix indicate the associated currency requirements. 

Table II shows the relationships from function nodes to resource nodes. The near-diagonal structure of this this matrix reflects that, in the current model, each information resource is linked to a single functional means (with the exception of synchrony functions). This assumption simplifies the simulation of currency requirements: when an information resources becomes outdated, only one function can be invoked for its update. In contrast, systems with multiple functional means per resource would require context-dependent selection mechanisms and more complex computational logic to be able to simulate the work model. Such extensions are beyond the scope of this case study.

\begin{table*}[]
    \centering
    \caption{Adjacency matrix of edges from resources to functions with associated currency values.}
    \label{tab:placeholder}
    \setlength{\tabcolsep}{1pt}
    \renewcommand{\arraystretch}{1}
    \begin{tabular}{r||c|c|c|c|c|c|c|c|c|c|c|c|c|c||c|c|c}
    \diagbox{Ends}{Means} 
    & \rotatebox{90}{\parbox{2.5cm}{\flushleft Observed Battery Level (OBL)}} 
    & \rotatebox{90}{\parbox{2.5cm}{\flushleft Camera Imagery (CIMG)}} 
    & \rotatebox{90}{\parbox{2.5cm}{\flushleft Observed Subsystem Temperatures (OST)}} 
    & \rotatebox{90}{\parbox{2.5cm}{\flushleft Observed Rover State (ORS)}} 
    & \rotatebox{90}{\parbox{2.5cm}{\flushleft Rover State (RS)}} 
    & \rotatebox{90}{\parbox{2.5cm}{\flushleft Goal Location (GL)}} 
    & \rotatebox{90}{\parbox{2.5cm}{\flushleft Planned Path (PP)}} 
    & \rotatebox{90}{\parbox{2.5cm}{\flushleft Next Waypoint (NWP)}} 
    & \rotatebox{90}{\parbox{2.5cm}{\flushleft Obstacle Locations (OL)}} 
    & \rotatebox{90}{\parbox{2.5cm}{\flushleft Terrain Map (TM)}} 
    & \rotatebox{90}{\parbox{2.5cm}{\flushleft Obstacle Size (OS)}} 
    & \rotatebox{90}{\parbox{2.5cm}{\flushleft Camera Angles (CA)}} 
    & \rotatebox{90}{\parbox{2.5cm}{\flushleft Goal Reached (GR)}} 
    & \rotatebox{90}{\parbox{2.5cm}{\flushleft Robot's Activity Status (RAS)}} 
    & \rotatebox{90}{\parbox{2.5cm}{\flushleft Directions for Robot Activity (DRA)}} 
    & \rotatebox{90}{\parbox{2.5cm}{\flushleft Awareness of Robot Activity (ARA)}} 
    & \rotatebox{90}{\parbox{2.5cm}{\flushleft Projections of Robot Activity (PRA)}} \\
    \hline
    \text{Battery Level Monitoring (BLM)}            &    &   &    &    & LT &    &   &    &    &    &    &    &    &    &    &     & \\
    \text{Temperature Monitoring (TMP)}              &    &   &    &    & LT &    &   &    &    &    &    &    &    &    &    &     & \\
    \text{Location and Attitude Assessment (LAA)}    &    & 1 &    &    &    &    &   &    &    & 30 &    & 1  &    &    & 60 &     & 120\\
    \text{Robot Path Planning (RPP)}                 & LT &   & 10 & LT &    & 30 &   &    & 10 &    & LT &    &    &    & 60 &     & 120\\
    \text{Next Waypoint Selection (NWS)}             &    &   &    &    &    &    & x &    &    &    &    &    &    &    &    &     &  \\
    \text{Obstacle Localization (OLL)}               &    & 0 &    &    &    &    &   &    &    &    &    & 1  &    &    & 60 &     & 120\\
    \text{Image Capture (IC)}                        &    &   &    &    & LT &    &   &    &    &    &    & LT &    &    &    &     &  \\
    \text{Robot Movement (RM)}                       &    &   &    &    &    &    &   & LT &    &    &    &    &    &    &    &     & \\
    \text{Camera Angle Movement (CAM)}               &    &   &    &    &    &    &   &    &    &    & LT &    &    &    &    &     & 120\\ \hline
    \text{Monitoring Robot Activity (MonRA)}        &    &   &    &    &    &    &   &    &    &    &    &    &    & LT &    &     & \\
    \text{Directing Robot Activity (DirRA)}         &    &   &    &    &    &    &   &    &    &    &    &    &    &    &    & 0   & 120\\
    \text{Projecting Robot Activity (ProjRA)}       &    &   &    &    &    &    &   &    &    &    &    &    &    &    &    & LT  & \\
    \end{tabular}
    
    \vspace{5pt}
    \parbox{0.9\textwidth}{\footnotesize \textit{Note:} $x$ is an independent variable in the computational experiment, with $x = \{10, 15, 20, \ldots, 70\}$. Goal Reached is a terminal variable representing task completion and serving as the simulation’s stopping criterion. LT is a Long Time which in the simulation defaulted to $10^6$.}
\end{table*}

\begin{table*}[]
    \centering
    \caption{Adjacency matrix of edges from functionsto resources.}
    \label{tab:placeholder-transposed}
    \setlength{\tabcolsep}{1pt}
    \renewcommand{\arraystretch}{1}
    \begin{tabular}{r||c|c|c|c|c|c|c|c|c||c|c|c}
    \diagbox{Ends}{Means} 
        & \rotatebox{90}{\parbox{2.3cm}{\flushleft Battery Level Monitoring (BLM)}} 
        & \rotatebox{90}{\parbox{2.3cm}{\flushleft Temperature Monitoring (TMP)}} 
        & \rotatebox{90}{\parbox{2.3cm}{\flushleft Location and Attitude Assessment (LAA)}} 
        & \rotatebox{90}{\parbox{2.3cm}{\flushleft Robot Path Planning (RPP)}} 
        & \rotatebox{90}{\parbox{2.3cm}{\flushleft Next Waypoint Selection (NWS)}} 
        & \rotatebox{90}{\parbox{2.3cm}{\flushleft Obstacle Localization (OLL)}} 
        & \rotatebox{90}{\parbox{2.3cm}{\flushleft Image Capture (IC)}} 
        & \rotatebox{90}{\parbox{2.3cm}{\flushleft Robot Movement (RM)}} 
        & \rotatebox{90}{\parbox{2.3cm}{\flushleft Camera Angle Movement (CAM)}} 
        & \rotatebox{90}{\parbox{2.3cm}{\flushleft Monitoring Robot Activity (MonRA)}} 
        & \rotatebox{90}{\parbox{2.3cm}{\flushleft Directing Robot Activity (DirRA)}} 
        & \rotatebox{90}{\parbox{2.3cm}{\flushleft Projecting Robot Activity (ProjRA)}} \\
    \hline
    \text{Terrain Map (TM)}                      &   &   &   &   &   &   &   &   &   &   &   &   \\
    \text{Obstacle Size (OS)}                    &   &   &   &   &   &   &   &   &   &   &   &   \\
    \text{Goal Location (GL)}                    &   &   &   &   &   &   &   &   &   &   &   &   \\
    \text{Observed Battery Level (OBL)}          & 1 &   &   &   &   &   &   &   &   &   &   &   \\
    \text{Observed Subsystem Temperatures (OST)} &   & 1 &   &   &   &   &   &   &   &   &   &   \\
    \text{Observed Rover State (ORS)}            &   &   & 1 &   &   &   &   &   &   &   &   &   \\
    \text{Planned Path (PP)}                     &   &   &   & 1 &   &   &   &   &   &   &   &   \\
    \text{Next Waypoint (NWP)}                   &   &   &   &   & 1 &   &   &   &   &   &   &   \\
    \text{Obstacle Locations (OL)}               &   &   &   &   &   & 1 &   &   &   &   &   &   \\
    \text{Camera Imagery (CIMG)}                 &   &   &   &   &   &   & 1 &   &   &   &   &   \\
    \text{Rover State (RS)}                      &   &   &   &   &   &   &   & 1 &   &   &   &   \\
    \text{Camera Angles (CA)}                    &   &   &   &   &   &   &   &   & 1 &   &   &   \\ 
    \text{Goal Reached (GR)}                     &   &   &   &   &   &   &   & 1 &   &   &   &   \\ \hline
    \text{Robot's Activity Status (RAS)}         &   &   & 1 & 1 &   & 1 &   &   &   &   &   &   \\
    \text{Awareness of Robot Activity (ARA)}     &   &   &   &   &   &   &   &   &   & 1 &   &   \\
    \text{Directions for Robot Activity (DRA)}   &   &   &   &   &   &   &   &   &   &   & 1 &   \\
    \text{Projections of Robot Activity (PRA)}   &   &   &   &   &   &   &   &   &   &   &   & 1 \\
    \end{tabular}

    \vspace{5pt}
    \parbox{0.9\textwidth}{\footnotesize \textit{Note:} Terrain Map, Obstacle Size, and Goal Location are initialized prior to the simulation and remain static throughout. They could be modeled as dynamic variables in extended versions of the framework, which is beyond the scope of this study.}
\end{table*}

% \bibliographystyle{IEEEtran}
% \bibliography{IEEEabrv,references}
 % argument is your BibTeX string definitions and bibliography database(s)
%\bibliography{IEEEabrv,../bib/paper}
%